\documentclass[a4paper,conference]{IEEEtran}
\usepackage{url}

\usepackage{caption}
\usepackage{subcaption}

\ifCLASSINFOpdf
  \usepackage[pdftex]{graphicx}
\else
\fi
\usepackage{multirow}

\usepackage{amsmath}
\usepackage{amssymb}
\usepackage{algorithm}
\usepackage{algpseudocode}

\usepackage{array}

\usepackage{fixltx2e}
\usepackage{stfloats}
\usepackage{verbatim}
\usepackage[symbol]{footmisc}

\begin{document}
%
\title{SegDenseNet: Iris Segmentation for Pre and Post Cataract Surgery}


\author{Aditya Lakra\footnotemark{*}, Pavani Tripathi\footnotemark{*}, Rohit Keshari, Mayank Vatsa, Richa Singh \\
IIIT-Delhi, India\\
Email: \{aditya13006, pavani13147, rohitk, mayank, rsingh\}@iiitd.ac.in}


\maketitle

\begin{abstract}
Cataract is caused due to various factors such as age, trauma, genetics, smoking and substance consumption, and radiation. It is one of the major common ophthalmic diseases worldwide which can potentially affect iris-based biometric systems. India, which hosts the largest biometrics project in the world, has about 8 million people undergoing cataract surgery annually. While existing research shows that cataract does not have a major impact on iris recognition, our observations suggest that the iris segmentation approaches are not well equipped to handle cataract or post cataract surgery cases. Therefore, failure in iris segmentation affects the overall recognition performance. This paper presents an efficient iris segmentation algorithm with variations due to cataract and post cataract surgery. The proposed algorithm, termed as SegDenseNet, is a deep learning algorithm based on DenseNets. The experiments on the IIITD Cataract database show that improving the segmentation enhances the identification by up to 25\% across different sensors and matchers.  
\end{abstract}

\renewcommand{\thefootnote}{\fnsymbol{footnote}}
\footnotetext[1]{Equal contribution by student authors}
\renewcommand{\thefootnote}{\arabic{footnote}}

%
\IEEEpeerreviewmaketitle

\section{Introduction}

Iris recognition is one of the most reliable technologies available for person identification. It has gained a lot of importance due to the high accuracies and availability of commercial and open source iris recognition softwares such as Iricore\footnote{\url{https://goo.gl/TLcyAs}}, VeriEye\footnote{\url{http://www.neurotechnology.com/verieye.html}}, MIRLIN\footnote{\url{https://www.fotonation.com/products/biometrics/iris-recognition/}}. Owing to high discriminability of iris patterns, several large-scale systems are utilizing iris recognition systems. For instance, Government of India's Aadhaar project \cite{uidai} has enrolled over 1.2 billion citizens in the system using three biometric modalities, viz. face, fingerprint, and iris. 

\begin{figure}[t]
\centering
\includegraphics[width=3.4in]{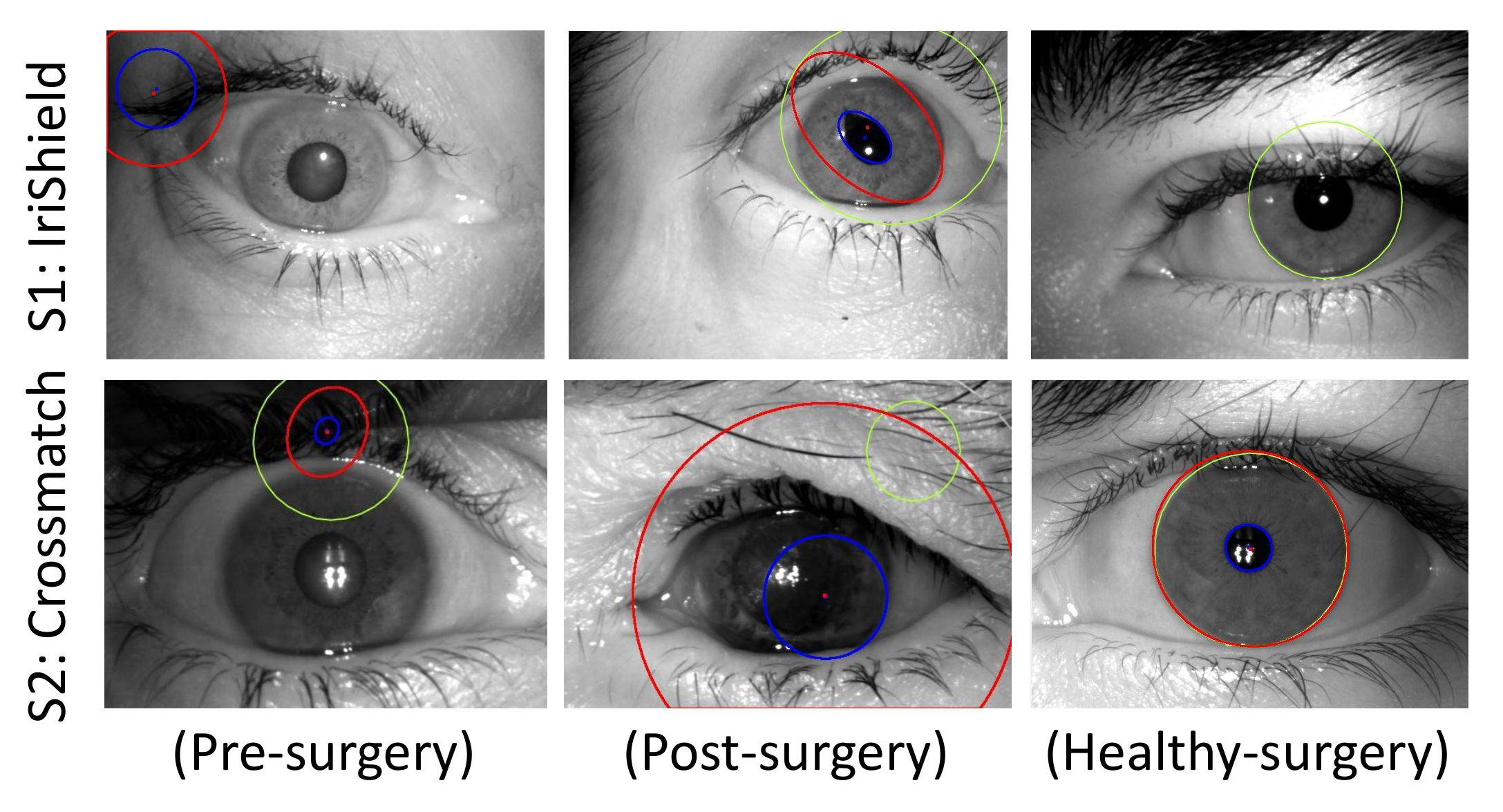}
\caption{Illustrating the challenges with iris segmentation in cataract affected eyes. Two commercial-of-the-shelf (COTS) systems have been used to segment the cataract affected iris images. Green and red \& blue circles represent segmentation using VeriEye and Matcher-1, respectively.}
\label{fig:fig_samples}
\vspace{-10pt}
\end{figure}

In such systems which include population of all age groups, several different kinds of challenges are encountered. For instance, elderly people have difficulty in opening the eyes and it is difficult to ensure that young kids keep the eyes stable so that it can be captured properly. Another important covariate faced by iris recognition systems is the presence of ocular diseases. Cataract is one of the several leading ophthalmic diseases in several countries. In 2010, more than 800,000 eyes in Germany \cite{Germany} and more than 81,500 eyes in Austria underwent cataract surgery \cite{Austria}. In India, it has been reported that the Aadhaar system will be host to approximately $8,000,000$ cataract patients undergoing surgery annually by 2020 \cite{Vision_2020}. This poses a challenge for the researchers in the biometrics field, since the state-of-the-art algorithms fail to give promising results on patients who have undergone cataract surgery \cite{Reliability_of_IR}. For instance, Fig. \ref{fig:fig_samples} shows the mis-classifications generated by state-of-the-art iris segmentation algorithms\footnote{Due to license restriction, we cannot name the Matcher-1.}. The current recommendations include that the individuals have to be re-enrolled into the system a few weeks after the surgery \cite{uidai, Reliability_of_IR}. This implies that approximately 8 million citizens will have to be re-enrolled annually by 2020. Since re-enrollment is a manually expensive task, this research explores an alternative strategy to improve the performance of iris recognition.

Researchers have made significant efforts in studying the effect of cataract surgery on the iris pattern and structure. Roizenblatt et al. \cite{Roizenblatt2004} were the first one to study the consequences of cataract surgery on iris pattern. Their hypothesis was that iris recognition systems fail due to changes such as depigmentation and localized iris atrophy, with loss of large areas of Fuchs’ crypts, circular, and radial furrows and pupil ovalization. However, Dhir et al. \cite{Cataract_influence} performed a study on a small dataset of 15 subjects and concluded that cataract surgery does not tamper the iris pattern. Rather, the dilation of pupil and specular reflections in the pupil region are the reasons why conventional circle-based algorithms and light-position-based pupil algorithms fail to achieve high accuracy. Further, Preethi and Jayanthi \cite{preethi2017som} observed that structural changes occur in iris. Thus, it is advisable to re-enroll the patient after surgery. Recently, Nigam et al. \cite{nigam2015ocular} showed that, in different ophthalmic diseases cases,  incorrect segmentation of iris region is a potential cause of reduced accuracy of state-of-the-art iris recognition systems. It is imperative to improve the existing segmentation algorithms to ensure accurate segmentation of the iris regions in pre and post cataract surgery images. 

\subsection{Literature Review}

Iris segmentation on images acquired under unconstrained environment have been explored by researchers. Zhao and Kumar \cite{Ajay2015} proposed a method which segments out the iris region based on the pupillary and limbic regions as well as by fitting a parabolic equation around the eyelids. Iris region has a lot of intensity, textural and structural information which can be used to for segmentation. Proenca \cite{colour_based} proposed a color based iris segmentation technique which classifies the iris pixels using a neural network. Tan and Kumar \cite{SVM_iris} used support vector machines (SVM) for classifying the Zernike moments extracted from around the pixels. These are few examples of pixel-based and boundary-based iris segmentation techniques. However, these algorithms are based on enormous amount of pre-knowledge of the dataset, as well as, pre and post-processing. Liu et al. \cite{IrisSeg_2016} proposed a deep learning based architecture for iris segmentation. The model gave state-of-the-art accuracy on CASIA v4-Distance and UBIRISv2 datasets and does not require post-processing. Recently, Jalilian et al. \cite{CNN_iris_latest} proposed linear and non-linear based domain adaptation methods for iris segmentation using fully convolutional network.

\subsection{Research Contributions}
To the best of authors' knowledge no research has been performed towards improving segmentation algorithms such that it can segment out the iris region from pre and post cataract surgery images. Developing an algorithm that can segment the iris region from eyes of cataract patients and those who have underwent cataract surgery, is the goal of this paper. This also leads to the elimination of the re-enrollment process. The major contributions of this research are:

\begin{enumerate}
  \item We propose a deep learning based algorithm which can segment the iris pattern from images of eyes taken pre and post cataract surgery.
  \item Since there is no pre and post cataract surgery database with annotated segmentation ground truth, we have generated ground truth masks for the IIITD Cataract Surgery Database using Adobe Photoshop\footnote{\url{http://www.adobe.com/in/products/photoshop.html}}. The database has images pertaining to 132 subjects and they are acquired using three different sensors, namely Crossmatch, IriShield, and Vista. These annotations are used for training and evaluating the performance of the proposed segmentation algorithms and they will also be released to the research community to promote research on this problem. 
\end{enumerate}


\begin{figure*}[t]
\centering
\includegraphics[width=7.2in]{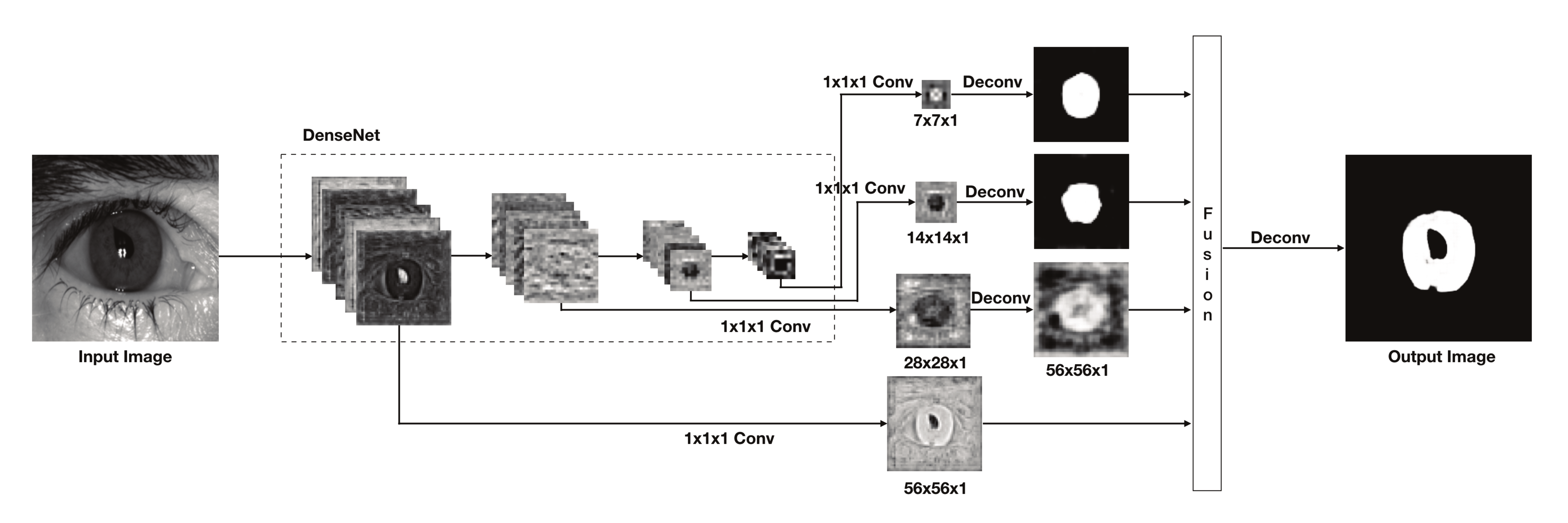}
\caption{Architecture of the proposed iris segmentation algorithm.}
\label{fig:fig_arch}
\end{figure*}

\section{SegDenseNet: Iris Segmentation Algorithm}

Convolutional neural networks have demonstrated to achieve very high accuracies in several recognition tasks. Inspired by its success, researchers have also explored their effectiveness for segmenting natural images. Long et al. \cite{FCN} modeled segmentation as a classification problem and designed Fully Convolutional Networks (FCN) for semantic segmentation. FCN is a special kind of neural network which only has convolution layers, pooling layers, and upsampling layers. It can input an arbitrary sized image and output a mask of corresponding size. 

Chen et al. \cite{ResNet_semantic_1} used ResNet \cite{ResNet} architecture and showed that with deep architectures, semantic segmentation accuracy can be improved significantly. Drozdzal et al. \cite{Tiramisu} proposed a deep learning based architecture for semantic segmentation using DenseNet \cite{densenet} and demonstrated that deep architecture indeed increases the segmentation accuracy.
However, iris segmentation is somewhat different from segmenting natural images. While semantic segmentation requires global as well as local information, iris segmentation requires very fine details such as inclusion of iris features, sharp boundary of the iris, and disposing the fine iris regions occluded by reflection, eye-lash and hair. To the best of our knowledge, no deep learning based algorithm exists for iris segmentation for cataract patients. During cataract, a cloudy white layer forms on the pupil, due to which state-of-the-art iris segmentation algorithms which use pixel intensity values to segment the iris fail to segment out the correct region of interest. In post cataract surgery, the specular reflections, morphological changes on the iris and pupil, increase the failure cases for existing methods. 

\begin{figure}[]
\centering
\includegraphics[width=3.5in]{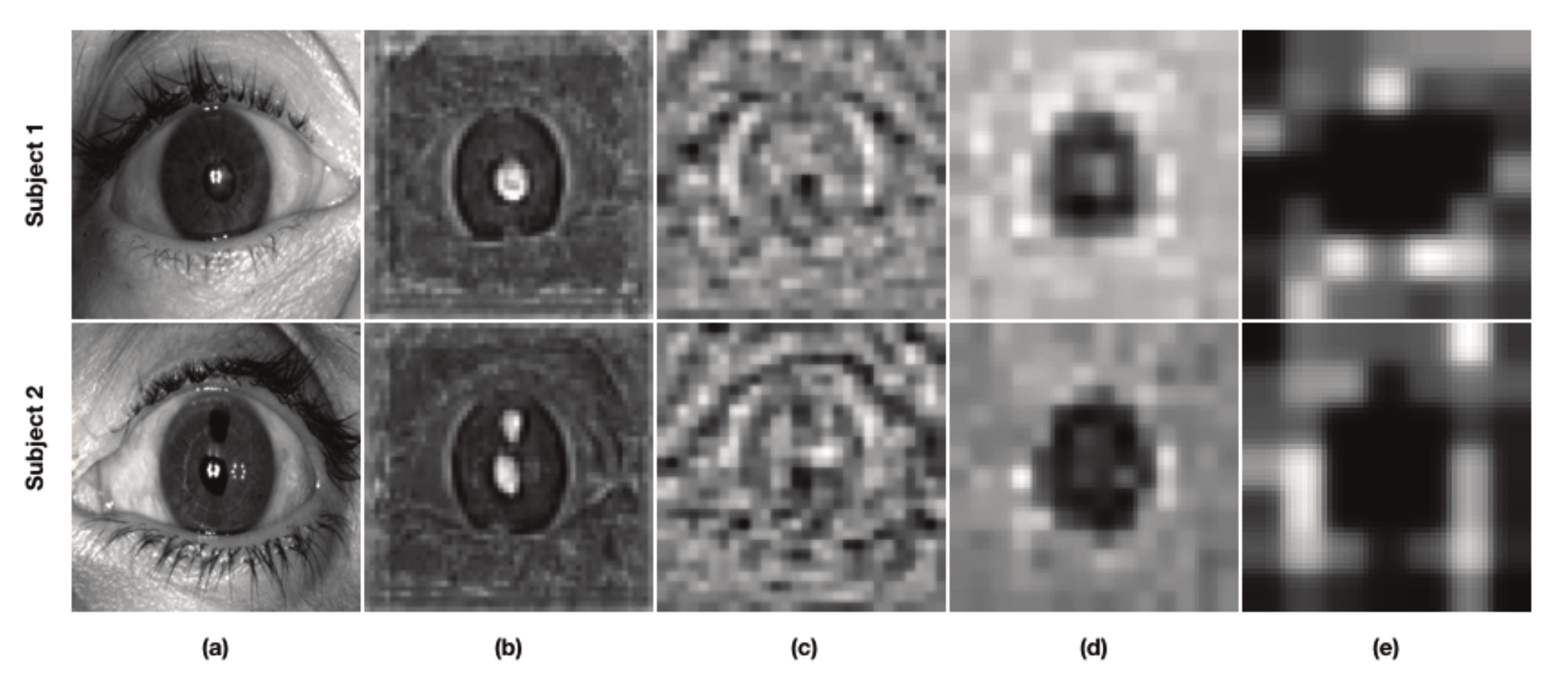}
\caption{Outputs of the four convolution blocks used for fusion. (a) Input image. Sample output of (b) first convolution block, (c) second convolution block, (d) third convolution block and (e) fourth convolution block}
\label{fig:fig_inter}
\end{figure}

Based on the idea behind FCN \cite{FCN}, we propose a deep learning architecture, SegDenseNet, for cataract affected iris segmentation which fuses the outputs from all the convolution blocks present in any deep learning architecture. In our implementation, we have used DenseNet \cite{densenet} with 121 convolution layers, having four convolution blocks. For fine prediction of the iris region, we fuse outputs from all the four dense blocks. In our architecture, the Fusion Layer performs a weighted sum operation on the four prediction maps as shown in equation~\ref{eq:wsum}. For our experiments all the intermediate feature maps have been given equal weight. The proposed architecture is presented in Fig. \ref{fig:fig_arch}. As illustrated in the figure, the number of channels in the mask is one because there is only one class: the iris region. Using convolution, the channel dimension of the outputs of all the four DenseNet blocks is reduced to 1, as shown in Eqs. \ref{eq:1}-\ref{eq:4}. These feature maps are then deconvolved as shown in Eqs. \ref{eq:1}-\ref{eq:3} to produce feature maps having coarse and fine details of the iris region. Finally, we fuse the four outputs (5) and deconvolve it to the size of the input image, to produce the desired mask (6). The sample outputs of the four convolutional blocks before they are converted to feature maps with 1 channel are visualized in Fig. \ref{fig:fig_inter}. 

\begin{equation}
I_{1} = Df_{n}(X'_{N'})   \otimes    Conv(1,1,1) \Psi Deconv(4,4,1)
\label{eq:1}
\end{equation}
\begin{equation}
I_{2} = Df_{n}(X'_{(N'-1)}) \otimes    Conv(1,1,1) \Psi  Deconv(8,8,1)
\label{eq:2}
\end{equation}
\begin{equation}
I_{3} = Df_{n}(X'_{(N'-2)}) \otimes    Conv(1,1,1) \Psi  Deconv(16,16,1)
\label{eq:3}
\end{equation}
\begin{equation}
I_{4} = Df_{n}(X'_{(N'-3)}) \otimes    Conv(1,1,1) 
\label{eq:4}
\end{equation}
\begin{equation}
I_{5} = w_{1}I_{1} + w_{2}I_{2} + w_{3}I_{3} + w_{4}I_{4}
\label{eq:wsum}
\end{equation}
\begin{equation}
M = Df_{n}(I_{5}) \Psi  Deconv(8,8,1)
\label{eq:final}
\end{equation}

Pseudo code of the proposed SegDenseNet is given in Algorithm \ref{SegDense}. The datasets A and B represent CASIA v4-Distance and IIITD Cataract Surgery Database respectively. $X_{N}^{tr}$ depicts the training set which contains the images and labels of dimension $224\times224$. The DenseNet-121 model pretrained on ImageNet dataset is modified according to our requirements. The images are first resized, $X_{N}^{tr'}$, and then augmented, $X_{N'}^{tr'}$. Post augmentation, we send the images and the labels in a batch for training as well as testing. We binarize the predicted masks and resize them to the size of the original image i.e. $640\times480$. The error is then computed between the ground truth mask, $M_{G}^{te'}$ and predicted mask, $M_{P}^{te'}$. The size of $X^{tr}$ and $M^{te'}$ is same. 

\begin{algorithm}[!t]
	\caption{Cataract Iris Segmentation}
	\label{SegDense}
	
	\begin{algorithmic}[1]
		\State \textbf{Database:} \{A, B\} 
        \State where:
        \State \hspace{4mm} A = CASIA v4-Distance dataset
        \State \hspace{4mm} B = IIITD Cataract Surgery Database 
		\State \textbf{Input:} $X_{N}^{tr} \leftarrow$ \{A, B\}; $X_{P}^{te} \leftarrow $\{B\}
		\State \textbf{Output:}	$M_{P}^{te}$
		\State \textbf{Model:}	$SDM \leftarrow SegDenseNet(DenseNet-121)$	
		\State $X_{N}^{tr'} \leftarrow resize(X_{N}^{tr})$
        \State $ X_{N'}^{tr'} \leftarrow augment(X_{N}^{tr'})$
		
		\For{Batch in  $X_{N'}^{tr'}$}
		\State train(SDM(Batch))
		\EndFor
		\For{Batch in  $X_{P}^{te}$}
		\State $M_{P}^{te} \leftarrow test(SDM(Batch))$
		\EndFor
		\State \textbf{Post-processing:} $M_{P}^{te'} \leftarrow resize(Binarize(M_{P}^{te})) $
		\State $error{=} \frac{1}{P\times m\times n} \sum_{i,j\epsilon (m,n)} M_G^{te'}(i,j)\bigoplus M_P^{te'}(i,j)$
	\end{algorithmic}
\end{algorithm}

\section{Database and Experimental Protocol}

The performance of the proposed iris segmentation algorithm is demonstrated on the IIITD Cataract Surgery database. Since the cataract database is small in size, CASIA v4-Distance database is used for training \textit{iris specific features}. 
In the following subsections, we describe the dataset details, experimental protocol, and the implementation details. 

\subsection{Datasets}
\textbf{CASIA v4-Distance} database ~\cite{CASIAv4} contains $2567$ face images pertaining to $142$ different subjects in the NIR spectrum. Since the images are captured from a distance, all the images have some level of noise. Each image captures a large portion of the whole face thus containing both the irises of the subject. After segmenting the eye region of these images, both right and left eye images are used for pre-training. 

\begin{figure}[!t]
\centering
\includegraphics[width=3in]{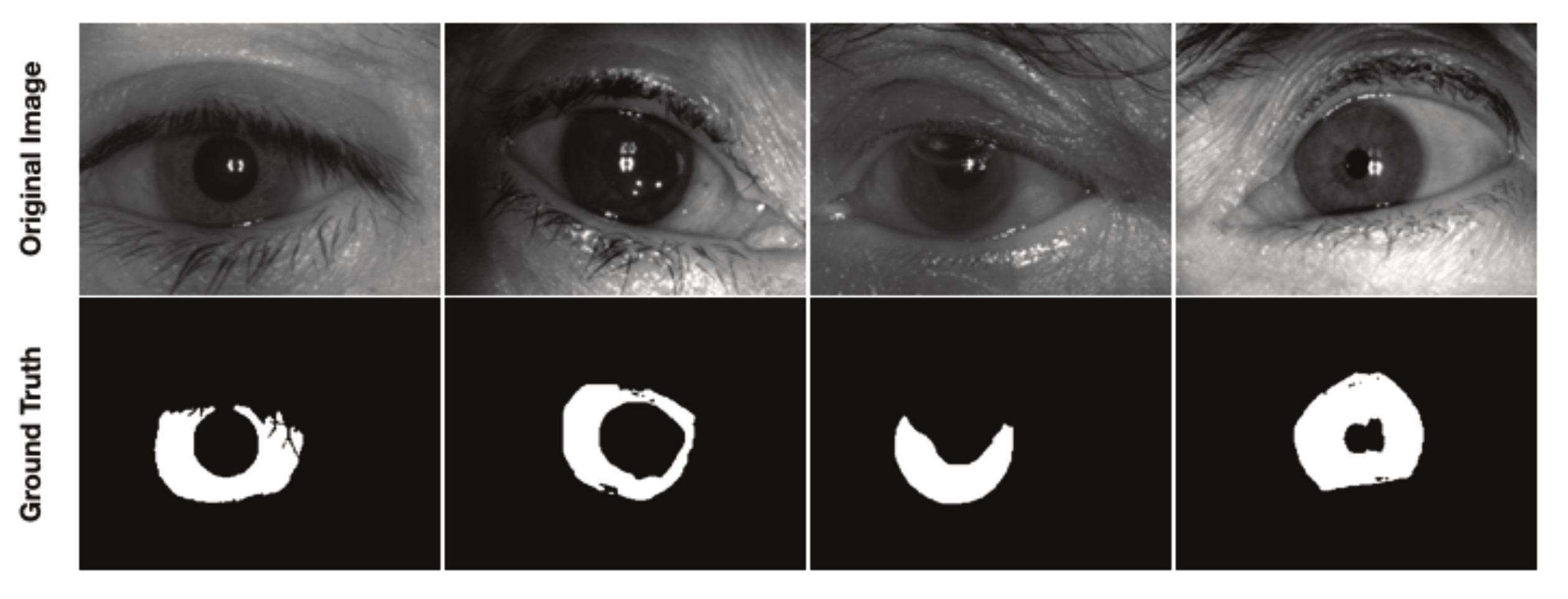}
\caption{Sample iris images and their manually generated ground truth masks.}
\label{fig:fig_sim}
\end{figure}

\textbf{IIITD Cataract Surgery Database\footnote{\url{http://iab-rubric.org/resources/ICSD.html}}} containing eye images with a total of $764$ post surgery and $764$ pre surgery images acquired from a total of $132$ subjects. However, the database only contains the class labels and not the segmentation annotations. To evaluate the segmentation performance, we manually annotated a total of $904$ ground truth masks. Fig. \ref{fig:fig_sim} shows the input images and their corresponding annotations. It can be observed visually how sharp, fine and accurate the ground truth masks are. These annotations will be publicly available to the research community \footnote{http://www.iab-rubric.org/resources.html}. 70\% database is utilized for training and remaining 30\% images (from unseen subjects) are used for computing the segmentation performance. The original size of the images is $640\times480$, however, the experiments are performed with images resized to $224\times224$. 

\begin{figure}[!t]
\centering
\includegraphics[width=3in]{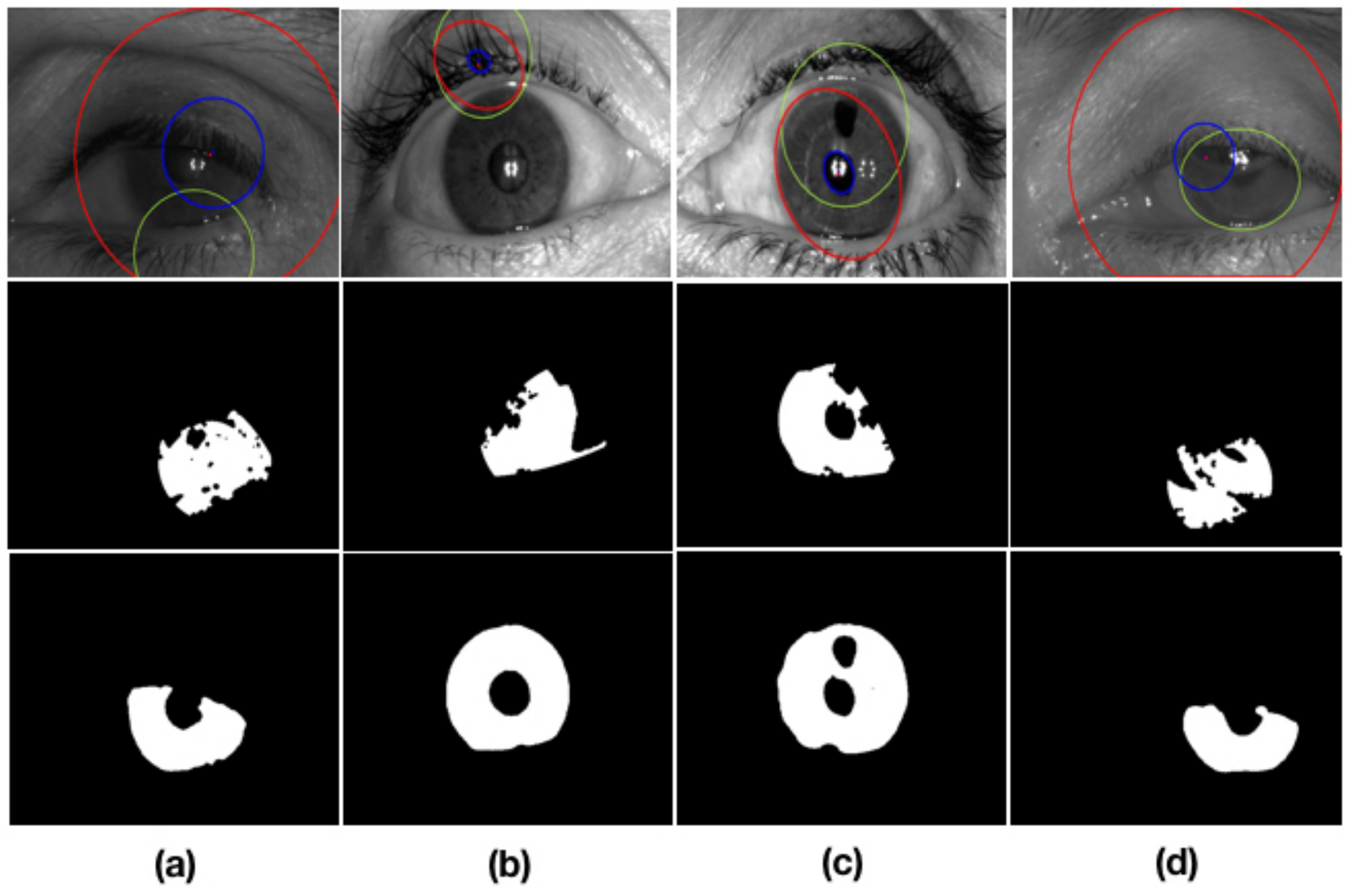}
\caption{Showcasing the results of iris segmentation on post (a)-(c) and pre (d)-(f) cataract surgery images. First row shows the segmentation results using COTS systems, second row of Zhao and Kumar \cite{Ajay2015}, and the last row pertains to the proposed algorithm.}
\label{fig:fig_pass_cases}
\end{figure}

\subsection{Implementation Details}
The proposed deep learning architecture has been implemented in Keras \cite{chollet2015} with Tensorflow backend. The DenseNet-121 model pretrained on imagenet is first trained using left and right healthy iris images from the CASIA v4-Distance dataset~\cite{CASIAv4}, followed by finetuning using IIITD Cataract Surgery Database. While training the model, the learning rate and momentum values are set as $0.001$ and $0.9$ respectively with SGD optimizer. The model was trained on NVIDIA GTX $1080$ti GPU for a total of $500$ epochs. Since deep learning architectures require large number of samples for training, images from both the datasets, CASIA v4-Distance \cite{CASIAv4} and IIITD Cataract Surgery Databases, are augmented. The augmentation is performed using contrast normalization and flip operations to increase the database size by $10$ times. For contrast normalization, we have used $4$ different contrast factors where contrast factor is the number of times by which the difference between a pixel value and the center value has to be multiplied.

\subsection{Fusion of Dense-blocks}
\begin{figure*}[!t]
\centering
\includegraphics[width=5.5in]{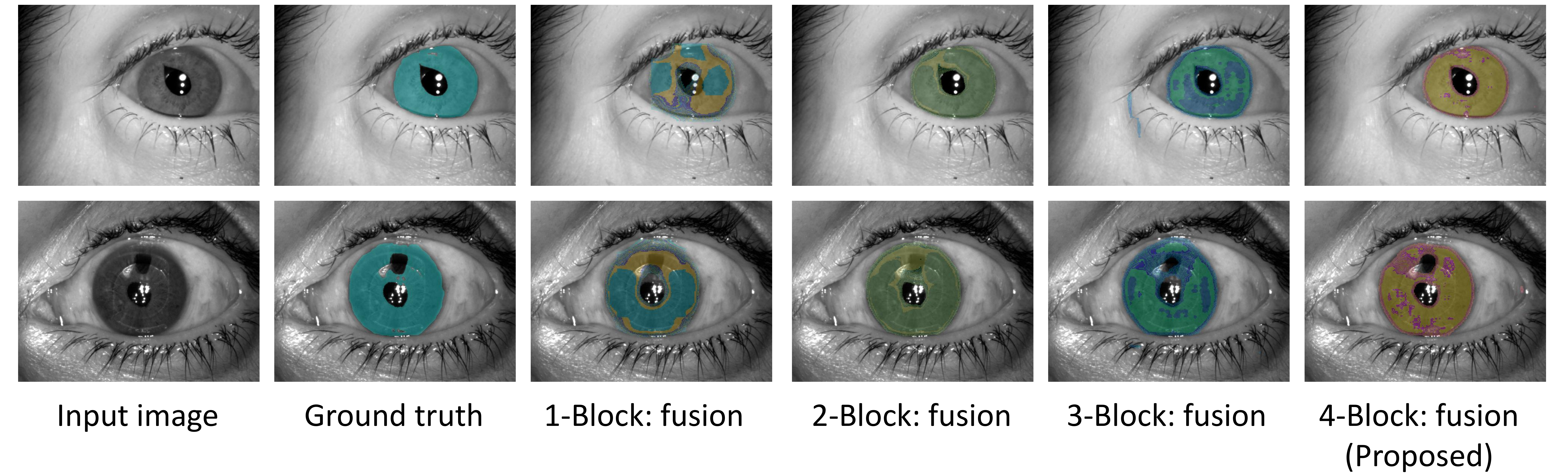}
\caption{Visually illustrating the effect of fusing different number of dense blocks on the segmentation performance (obtained masks are superimposed on the original image). The two different colours in each of the masks represent high and low confidence level.}
\label{fig:fig_confidence}
\end{figure*}

In our implementation, we have used DenseNet-121 which has four convolution blocks. Through visualization of masks, as depicted in Fig. \ref{fig:fig_confidence} it can be inferred that fusion of four blocks gives the best result. The two cases presented in the Fig. \ref{fig:fig_confidence} showcase the morphological changes that may occur post cataract surgery. The first row represents a case of pupil rupture. The protrusion in the pupil is completely segmented out only by the proposed method. In the case of the patient in second row, the surgery has resulted in a puncture inside the iris region. This deformity is completely segmented as non-iris region only when four Dense-blocks are fused. While in the other cases, some part of the puncture is segmented as iris region with low confidence. 

\section{Experimental Results and Analysis}

The performance of the proposed algorithm is evaluated in terms of both segmentation and matching performance. For segmentation, the results are compared with Zhao and Kumar \cite{Ajay2015}, while the impact of the segmentation algorithm on matching is demonstrated using a commercial system and the deep learning based algorithm by Zhao and Kumar \cite{dr_ajay_recog_2017}.


\vspace{6pt}
\noindent \textbf{Segmentation Performance}: The performance of the segmentation algorithm is measured in terms of metric from NICE-I competition \cite{NICE}. The average classification error rate is calculated using equation \ref{eq:metric}.

\begin{equation}
Error = \frac{1}{N\times m\times n} \sum_{i,j = 1}^{m,n} M_G^{te'}(i,j)\oplus M_P^{te'}(i,j)
\label{eq:metric}
\end{equation}

\noindent In equation \ref{eq:metric}, $N$, $m$ and $n$ denote the total number of test samples, height of the mask, and width of the mask respectively. The logical exclusive-or operator calculates the proportion of the correspondent disagreeing pixels.

Fig. \ref{fig:fig_pass_cases} shows failed cases of COTS system. The results obtained using state-of-the-art iris segmentation technique proposed by Zhao and Kumar \cite{Ajay2015} are also shown in the figure along with the iris masks obtained using SegDenseNet. It should be noted that Zhao and Kumar's method involves parameter tuning. We have tried to fine-tune the parameters to the best of our ability. It achieves average segmentation error of 6.28\% after parameter tuning. The proposed deep learning based segmentation algorithm yields 0.98\% average segmentation error. The average segmentation error for both the algorithms are compiled in Table \ref{table_seg}. 

\begin{table}[!t]
\renewcommand{\arraystretch}{1.3}
\caption{Average segmentation error rates on IIITD Cataract Surgery Database.}
\label{table_seg}
\centering
\begin{tabular}{|c|c|}
\hline
\textbf{Method} & \textbf{Error(\%)}\\
\hline
Proposed Method & 0.98\\
\hline
 Zhao and Kumar \cite{Ajay2015} & 6.28\\
\hline
\end{tabular}
\end{table}

\begin{table*}[!t]
\centering
\caption{Comparing verification accuracies at 0.1\% false accept rate on the IIITD Cataract Surgery database.}
\label{table_recog}
\renewcommand{\arraystretch}{1.1}
\begin{tabular}{|l|l|c|c|c|c|}
\hline
\textbf{Sensor}             & \textbf{Experiment}                            & \textbf{Subjects}   & \textbf{Pre-Pre (\%)} & \textbf{Post-Post (\%)} & \textbf{Pre-Post (\%)} \\ \hline
\multirow{6}{*}{CrossMatch} & Matcher-1                                      & \multirow{6}{*}{44} & 97.1                  & 76.5                    & 68.2                   \\ \cline{2-2} \cline{4-6} 
                            & Matcher-1 + SegDenseNet                        &                     & 99.3                  & 84.7                    & 92.1                   \\ \cline{2-2} \cline{4-6} 
                            & Matcher-1 + SegDenseNet + Post-processing      &                     & 99.5                  & 86.8                    & \textbf{95.6}                   \\ \cline{2-2} \cline{4-6}  
                            & Zhao and Kumar \cite{dr_ajay_recog_2017,Ajay2015}                                  &                     & 43.1                  & 64.1                    & 25.2                   \\ \cline{2-2} \cline{4-6} 
                            & Zhao and Kumar \cite{dr_ajay_recog_2017} + SegDenseNet                   &                     & 80.7                  & 78.0                    & 56.8                   \\ \cline{2-2} \cline{4-6} 
                            & Zhao and Kumar \cite{dr_ajay_recog_2017} + SegDenseNet + Post-processing &                     & 89.8                     & 70.7                        & 61.2                       \\ \hline\hline
\multirow{6}{*}{IriShield}  & Matcher-1                                      & \multirow{6}{*}{39} & 97.4                  & 81.1                    & 73.7                   \\ \cline{2-2} \cline{4-6} 
                            & Matcher-1 + SegDenseNet                        &                     & 99.6                  & 92.5                    & 95.5                   \\ \cline{2-2} \cline{4-6} 
                            & Matcher-1 + SegDenseNet + Post-processing      &                     & 99.6                  & 93.2                    & \textbf{96.7}                   \\ \cline{2-2} \cline{4-6} 
                            & Zhao and Kumar \cite{dr_ajay_recog_2017,Ajay2015}                                 &                     & 78.6                  & 89.7                    & 56.9                   \\ \cline{2-2} \cline{4-6} 
                            & Zhao and Kumar \cite{dr_ajay_recog_2017} + SegDenseNet                   &                     & 95.3                  & 85.7                    & 72.4                   \\ \cline{2-2} \cline{4-6} 
                            & Zhao and Kumar \cite{dr_ajay_recog_2017} + SegDenseNet + Post-processing &                     & 95.2                      & 83.1                        & 72.0                       \\ \hline\hline
\multirow{6}{*}{Vista}      & Matcher-1                                    & \multirow{6}{*}{27} & 96.8                  & 83.7                    & 75.0                     \\ \cline{2-2} \cline{4-6} 
                            & Matcher-1 + SegDenseNet                        &                     & 99.2                  & 92.9                    & 90.9                   \\ \cline{2-2} \cline{4-6} 
                            & Matcher-1 + SegDenseNet + Post-processing      &                     & 99.4                  & 94.1                    & \textbf{95.4}                   \\ \cline{2-2} \cline{4-6} 
                            & Zhao and Kumar \cite{dr_ajay_recog_2017,Ajay2015}                                 &                     & 3.4                   & 5.4                    & 0.7                    \\ \cline{2-2} \cline{4-6} 
                            & Zhao and Kumar \cite{dr_ajay_recog_2017} + SegDenseNet                   &                     & 91.4                  & 69.0                    & 43.1                   \\ \cline{2-2} \cline{4-6} 
                            & Zhao and Kumar \cite{dr_ajay_recog_2017} + SegDenseNet + Post-processing &                     & 90.1                      & 75.5                        & 44.3                       \\ \hline
\end{tabular}
\end{table*}

\vspace{6pt}
\noindent \textbf{Recognition Performance}: For calculating the iris recognition accuracy, we have used a state-of-the-art commercial software (Matcher I\footnote{Due to the license agreement with the manufacturer, the name of the matcher is kept anonymous.}) and Zhao and Kumar's \cite{dr_ajay_recog_2017} state-of-the-art deep learning based iris recognition. Table \ref{table_recog} summarizes the recognition accuracies obtained with and without the proposed segmentation algorithm, SegDenseNet.

\begin{figure*}[h]
\centering
\captionsetup[subfigure]{labelformat=empty}
\begin{subfigure}[b]{0.3\textwidth}
	\centering
	\includegraphics[width=\linewidth]{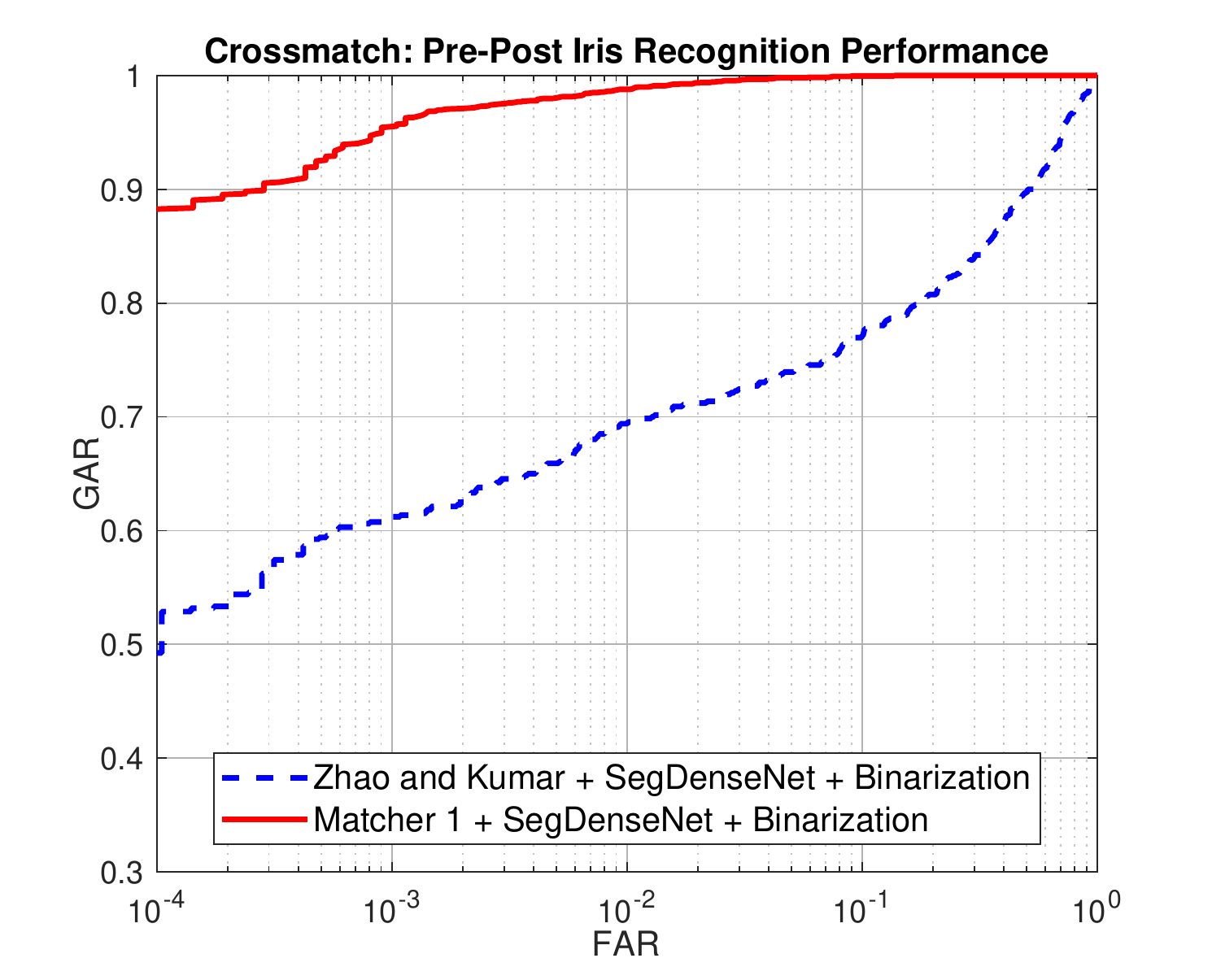}
	
	\label{fig:fig1}
\end{subfigure}
\qquad
\begin{subfigure}[b]{0.3\textwidth}
	\centering
	\includegraphics[width=\linewidth]{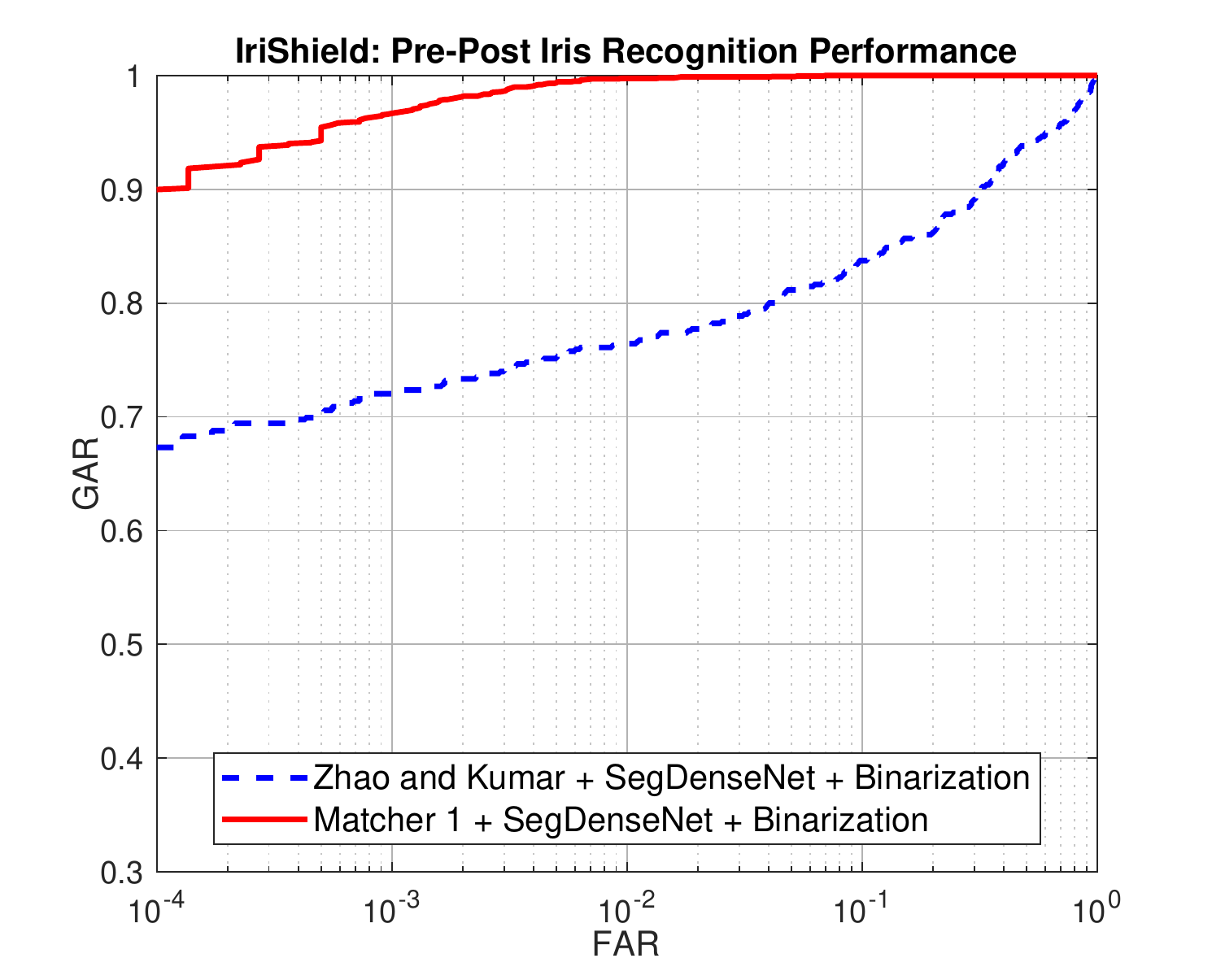}
	
	\label{fig:fig2}
\end{subfigure}
\qquad
\begin{subfigure}[b]{0.3\textwidth}
	\centering
	\includegraphics[width=\linewidth]{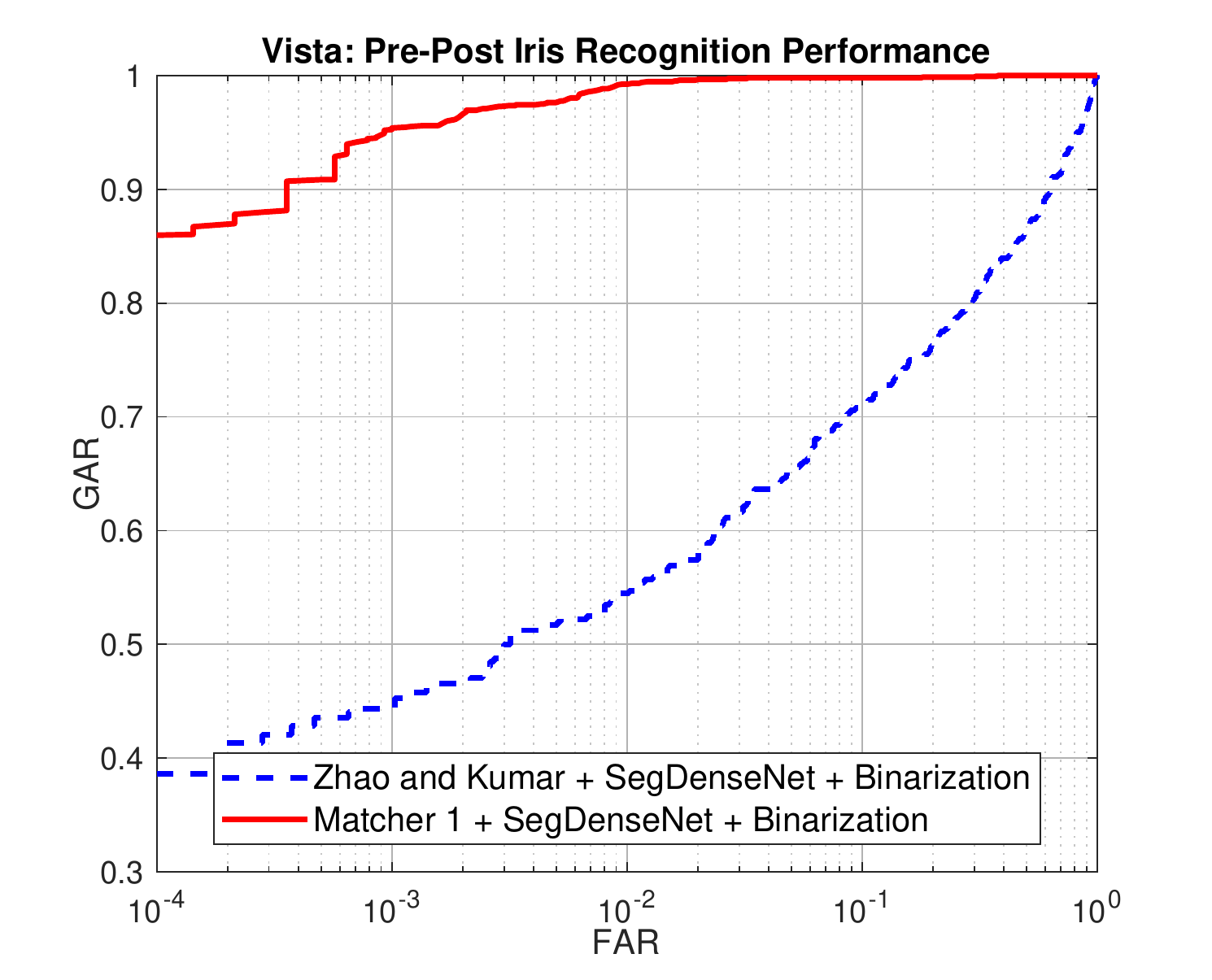}
	
	\label{fig:fig2}
\end{subfigure}
\caption{ROC curves for the recognition experiments performed on pre-post cataract surgery.}
\label{ROC}
\end{figure*}

The matching is performed amongst images captured before surgery (Pre-Pre), after surgery (Post-Post), and between images captured before and after surgery (Pre-Post). The recognition accuracies are summarized in Table \ref{table_recog}. It can be observed that there is a significant rise in the accuracies when SegDenseNet is used. For instance, with the COTS system recognition pipeline, incorporating SegDenseNet improved the accuracies by 2.4\%, 2.2\%, and 2.6\%, in case of Pre-Pre matching while using the three sensors, namely Crossmatch, IriShield and Vista, respectively. In case of Post-Post we have achieved 10.3\%, 12.1\% and 10.4\% increase in accuracies and in Pre-Post case the accuracies are improved by 27.4\%, 23.0\% and 20.4\% for the three sensors, respectively. Similar trend in the results are observed when Zhao and Kumar's \cite{dr_ajay_recog_2017} recognition pipeline was used. However, the accuracies were not as high as those obtained using COTS system. This could be because none of the publicly available models are trained using the cataract dataset. Also, due to failure of process 6 images were removed from the testing set. The ROC curves in Fig. \ref{ROC} compare the recognition accuracies obtained by Matcher-1 and Zhao and Kumar's \cite{dr_ajay_recog_2017} recognition pipeline when they are used with our proposed segmentation technique, SegDenseNet.  

\begin{figure}[!t]
\centering
\includegraphics[width=3.25in]{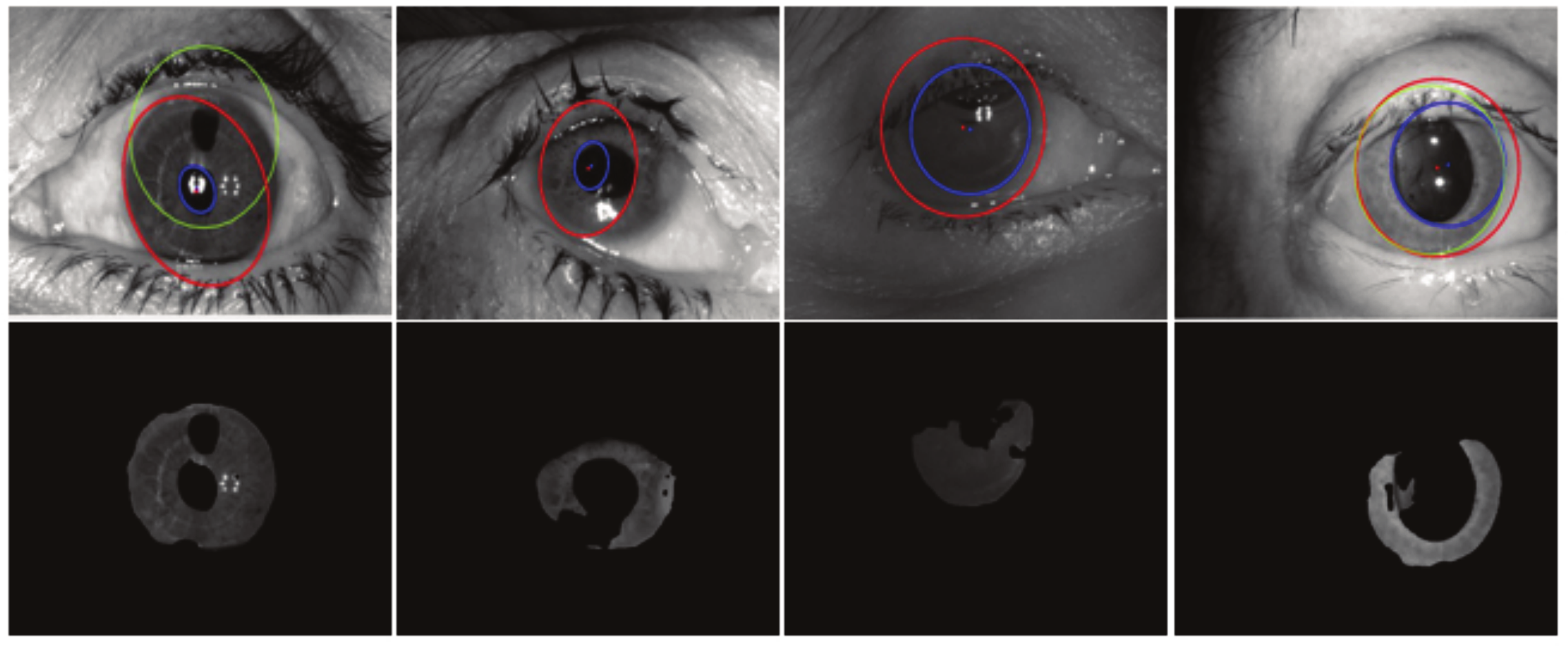}
\caption{Sample cases where COTS as well as the proposed method failed to segment the iris perfectly. Green, Red \& Blue circles represent output of VeriEye and Matcher-1 respectively.}
\label{fig:fig_fail_cases}
\end{figure}

Figure \ref{fig:fig_fail_cases} shows exemplars of images where the proposed algorithm and commercial SDKs failed to correctly segment out the iris region. On visually observing the images, we concluded that the failure happens due to bubbles in eyes. Also, in most of the failed cases, iris images are present at off-angle while there are very few training samples with off-angle iris region and/or bubbles. It is our assertion that if the model is trained with more number of such data samples, then the model should be able to successfully segment the iris region. 

\section{Conclusion}
Cataract and it's post-operative complications can cause iris recognition algorithms to fail, particularly at the segmentation stage. Particularly, irregular pupil shape and significant specular reflections after surgery lead to segmentation failures. Therefore, this research focuses on designing a segmentation algorithm for extracting iris region from eyes with cataract and post cataract surgery. We propose a deep learning based segmentation algorithm, named as SegDenseNet, that utilizes four dense blocks to learn model-specific iris shape even in presence of irregularities. Segmentation results on IIITD Cataract Surgery Database show that the proposed deep learning based segmentation algorithm outperforms existing segmentation algorithms for cataract affected eyes. The matching results also demonstrate the importance of improving the segmentation algorithms on the cataract and post-operative cases.



\bibliographystyle{IEEEtran}
\bibliography{IEEEexample}

\end{document}